\pdfoutput=1
\documentclass{article}
\usepackage[final,nonatbib]{nips_2018}

\usepackage[utf8]{inputenc} % allow utf-8 input
\usepackage[T1]{fontenc}    % use 8-bit T1 fonts
\usepackage{hyperref}       % hyperlinks
\usepackage{url}            % simple URL typesetting
\usepackage{booktabs}       % professional-quality tables
\usepackage{amsfonts}       % blackboard math symbols
\usepackage{nicefrac}       % compact symbols for 1/2, etc.
\usepackage{microtype}      % microtypography
\usepackage[draft=true]{minted}
\usepackage{graphicx}

\usepackage{amsmath}
\usepackage{amssymb}
\usepackage{latexsym}
\usepackage{ebproof}
\usepackage{listings}

\newcommand{\src}{\lstinline[mathescape, keepspaces]}
\newcommand{\msrc}[1]{\mbox{\src!#1!}}

\newcommand{\eval}{\ensuremath{$\,\reflectbox{$\leadsto$}\:$}}

\title{Composing Modeling and Inference Operations\\ with Probabilistic Program Combinators}

\author{
  Eli Sennesh\\
  Northeastern University\\
  Boston, MA\\
  \texttt{esennesh@ccis.neu.edu} \\
  \And
  Adam \'Scibior \\
  University of Cambridge and MPI T\"ubingen \\
  Cambridge, United Kingdom\\
  \texttt{ams240@cam.ac.uk} \\
  \And 
  Hao Wu\\
  Northeastern University \\
  Boston, MA\\
  \texttt{haowu@ccis.neu.edu} \\
  \And 
  Jan-Willem van de Meent\\
  Northeastern University \\
  Boston, MA\\
  \texttt{jwvdm@ccis.neu.edu} \\
}

\begin{document}
%\nipsfinalcopy

\maketitle

\begin{abstract}
Probabilistic programs with dynamic computation graphs can define measures over sample spaces with unbounded dimensionality, which constitute
programmatic analogues to Bayesian nonparametrics. Owing to the generality of this model class, inference relies on ``black-box'' Monte Carlo methods that are often not able to  take advantage of conditional independence and exchangeability, which have historically been the cornerstones of efficient inference. We here seek to develop a ``middle ground'' between probabilistic models with fully dynamic and fully static computation graphs. To this end, we introduce a \emph{combinator} library for the Probabilistic Torch framework. Combinators are functions that accept models and return transformed models. We assume that models are dynamic, but that model composition is static, in the sense that combinator application takes place prior to evaluating the model on data. Combinators provide primitives for both model and inference composition. Model combinators take the form of classic functional programming constructs such as \texttt{map} and \texttt{reduce}. These constructs define a computation graph at a coarsened level of representation, in which nodes correspond to models, rather than individual variables. Inference combinators implement operations such as importance resampling and application of a transition kernel, which alter the evaluation strategy for a model whilst preserving \emph{proper weighting}. Owing to this property, models defined using combinators can be trained using stochastic methods that optimize either variational or wake-sleep style objectives. As a validation of this principle, we use combinators to implement black box inference for hidden Markov models.
\end{abstract}

\section{Introduction}

Bayesian nonparametric models have traditionally leveraged exchangeability in order to define predictive distributions that marginalize over an unbounded number of degrees of freedom. In recent years, the field of probabilistic programming has explored a different (yet related) class of models. A probabilistic program can be thought of as a stochastic simulator that is conditioned on observed variables. 
When the probabilistic programming language supports recursion, probabilistic programs can define priors that sample from models with an unbounded number of random variables, providing a programmatic alternative to classic Bayesian nonparametric models. 
 
A probabilistic program must support two operations. First, it must be possible to generate samples by evaluating the program. In general, any (halting) evaluation instantiates some finite set of random variables, whose values are referred to as a \emph{trace}. The second operation that must be implemented is the evaluation of the unnormalized density function of a program for any trace. These operations can be formalized in two equivalent forms of  denotational semantics in which a program $f$ with inputs $y$ either evaluates to an unnormalized measure $\gamma_f(x \mid y)$, or a weighted sample $x,w \eval f(y)$ \cite{scibior2017denotational,Scibior:2018:FPM:3243631.3236778}. Inference seeks to characterize the target density $\pi_f(x \mid y) = \gamma_f(x \mid y) / Z_f(y)$. As in other inference problems, the integral $Z_f(y) = \int \gamma_f(x \mid y) \, dx$ is typically intractable, and is typically approximated using Monte Carlo methods.

In both nonparametric models and general probabilistic programs, we can significantly improve the performance
of approximate Bayesian inference by imposing some \emph{a priori} assumptions about the graphical form of the joint
distribution to be conditioned on our observations.  For example, in some models we can alternate between updates to
local and global plates of variables. In a Hidden Markov Model (HMM), for instance, we can predict transition probabilities
from state sequences, or vice versa.  Research in probabilistic programming has traditionally emphasized the development
of assumption-free inference methods.  To address model-specific inference optimizations, we develop abstractions to modularly and compositionally specify models and inference strategies.

\section{Model and Inference Composition}

In recent years, there have been a number of efforts to develop specialized inference methods for probabilistic programming. The Venture \cite{mansinghka2014venture} platform provides primitives for inference programming that can act on subsets of variables in an execution trace. 
%This allows a user to write a Gibbs sampler that iterates between, say, MCMC updates for a global variable and conditional SMC updates for local variables. 
There has also been work to formalize notions of valid inference composition. The Hakaru language \cite{narayanan2016probabilistic} frames inference as program
transformations, which can be composed so as to preserve a measure-theoretic denotation \cite{zinkov2017composing}. Work by Scibior et
al.~\cite{scibior2017denotational} defines measure-theoretic validity criteria for compositional inference transformations.

Models in Probabilistic Torch are written in Python and can make use of $\tt{if}$ expressions, loops, and other control flow constructs. Models can dynamically instantiate random variables in a data-dependent manner, although the computation graph becomes difficult to analyze statically\cite{schulman2015gradient}. To reason without static guarantees, we postulate these requirements for model and inference composition:

\textbf{1. Composition is static, evaluation is dynamic}. A model is statically composed from other models, while each evaluation based on data generates a unique trace. In our later HMM example, we can compose a model that samples global parameters with a model for a sequence of variably many states and observations.  Evaluations which traverse the same control-flow path while sampling different random values yield traces we can use as samples from the same distribution.

\textbf{2. Inference operations preserve proper weights.} A program $f$ defines a measure
$\gamma_f(x\mid y)$, which may be unnormalized, conditioned on some set of inputs $y$. This measure can represent
a prior distribution, or a distribution that is conditioned using observed variables or factors.  We here assume that valid evaluation strategies for a program yield \emph{properly
weighted} \cite{naesseth2015nested} samples $(X, W)$ such that, for all measurable functions $h$,
\begin{align}
    \mathbb{E}[h(X) W] &= \textstyle\int  h(x) \, \gamma_f(x \mid y) \: dx.
\end{align}
This property implies that $\mathbb{E}[W] = Z(y)$, i.e.~the weight $W$ is an unbiased estimator of the normalizer.  A default evaluation strategy that satisfies this assumption is likelihood weighting,
in which samples are proposed from the program prior and conditioning operations define an importance weight.

We require that any inference combinator must preserve proper weighting. Operations that satisfy this requirement include importance sampling, importance resampling, Sequential Monte Carlo (SMC), and application of a transition kernel. It follows that any composition of these operations also preserves proper weighting, resulting in an inference strategy that is properly weighted by construction\cite{scibior2017denotational}.

\section{Model Combinators}

A model is a stochastic computation that returns a properly weighted sample.  Model evaluation produces a trace, an
object that holds values and densities for the set of random variables instantiated during a particular evaluation of
the model.  Traces can be conditioned on other traces to implement proposals.
Combinators accept models as inputs and return a model.

Table~\ref{tab:big_step_modeling} shows a number of combinators corresponding to functional programming constructs, with
their semantics.  In addition to those for which we give the semantics, we have also implemented \src{reduce} as a basic
folding combinator, and such common model families as \src{ssm} (state-space model), \src{mixture}, and \src{hmm} (a
hidden Markov model).  Of particular note is that we can give semantics for higher-order stochastic functions built
using \src{partial} and \src{compose}.

\begin{table}
  \centering 
  \begin{tabular}{c}
      % \toprule 
  \toprule
  \textbf{Model Combinators} \\
  \midrule
  {\begin{prooftree}
      \hypo{ (x, w) \eval \msrc{$f$($y$)} }
      \hypo{ (x', w') \eval \msrc{$g$($x$)} }
      \infer2{ (x', w \cdot w')\eval \msrc{compose($f$,$g$)($y$)}}
  \end{prooftree}} \\[1.75em]
  %\hline
  {\begin{prooftree}
      \hypo{(x, w) \eval \msrc{$f$($y_1$,$y_2$)}}
      \infer1{ (x, w) \eval \msrc{partial($f$,$y_1$)($y_2$)} }
  \end{prooftree}} \\[1.75em]
  %\hline
  {\begin{prooftree}
      \hypo{ (x_n, w_n)\eval f(y_n)\, \text{~~for~} n = 1, \ldots, N} 
      \infer1{ ((x_1,\ldots,x_N), \prod_{n=1}^{N} w_n)\eval \msrc{map($f$,$(y_1, \ldots, y_N)$)}}
  \end{prooftree}} \\[1.25em]
  \bottomrule
  \end{tabular}
  % \begin{tabular}{c}
  %       % \toprule 
  %   \toprule
  %   \textbf{Inference Combinators} \\
  %   \midrule
  %   {\begin{prooftree}
  %       \hypo{(x,w)\eval\tt{g(y)}} 
  %       \hypo{w'=\gamma_f(x \mid y) / w}
  %       \infer2{ (x,w')\eval {\tt importance}(f, g)(y) }
  %   \end{prooftree}} \\[1.75em]
  %   %\hline
  %   {\begin{prooftree}
  %       \hypo{(x^k, w^k) \eval f(y)} 
  %       %\hypo{\bar{w} = \frac{1}{K} \sum_{k=1}^K w^k}
  %       %\hypo{\alpha^k \sim \text{Cat}\left(\frac{w^1}{K\bar{w}}, ..., \frac{w^K}{K\bar{w}}\right)}
  %       \hypo{\alpha^k \sim \text{Cat}\left(\frac{w^1}{\sum_{k=1}^K w^k}, ..., \frac{w^K}{\sum_{k=1}^K w^k}\right)}
  %       \infer2{ (x^{\alpha^k}, \frac{1}{K} \sum_{k=1}^K w^k)\eval {\tt resample}(f, k)(y) }
  %   \end{prooftree}} \\[1.75em]
  %   %\hline
  %   {\begin{prooftree}
  %       \hypo{(x, w)\eval f(y)} 
  %       \hypo{x'\sim q_{\tt{g}}(x'\mid x)}
  %       \hypo{w' = \frac{\gamma_{f}(x'\mid y)q_{g}(x\mid x')}{\gamma_{f}(x\mid y)q_{g}(x'\mid x)}w}
  %       \infer3{ (x',w')\eval {\tt move}(f,g)(y) }
  %   \end{prooftree}} \\[1.25em]
  %   \bottomrule
  % \end{tabular}
  \vspace{0.5em}
  \caption{Big-step semantics for model combinators.}
  \label{tab:big_step_modeling}
\end{table}

\begin{table}
  \centering 
  \begin{tabular}{c}
        % \toprule 
    \toprule
    \textbf{Inference Combinators} \\
    \midrule
    {\begin{prooftree}
        \hypo{(x,w)\eval g(y)} 
        \hypo{w'=\gamma_f(x \mid y) / w}
        \infer2{ (x,w')\eval  \msrc{importance($f$,$g$)($y$)} }
    \end{prooftree}} \\[1.5em]
    %\hline
    {\begin{prooftree}
        \hypo{(x^k, w^k) \eval f(y)} 
        %\hypo{\bar{w} = \frac{1}{K} \sum_{k=1}^K w^k}
        %\hypo{\alpha^k \sim \text{Cat}\left(\frac{w^1}{K\bar{w}}, ..., \frac{w^K}{K\bar{w}}\right)}
        \hypo{\alpha^k \sim \text{Cat}\left(\frac{w^1}{\sum_{k=1}^K w^k}, ..., \frac{w^K}{\sum_{k=1}^K w^k}\right)}
        \infer2{ (x^{\alpha^k}, \frac{1}{K} \sum_{k=1}^K w^k)\eval \msrc{resample($f$, $K$)($y$)} }
    \end{prooftree}} \\[2.0em]
    %\hline
    {\begin{prooftree}
        \hypo{(x, w)\eval f(y)} 
        \hypo{x'\sim q_{g}(x'\mid x)}
        \hypo{w' = \frac{\gamma_{f}(x'\mid y)q_{g}(x\mid x')}{\gamma_{f}(x\mid y)q_{g}(x'\mid x)}w}
        \infer3{ (x',w')\eval \msrc{move($f$,$g$)($y$)} }
    \end{prooftree}} \\[1.25em]
    \bottomrule
  \end{tabular}
  \vspace{0.5em}
  \caption{Big-step semantics for inference combinators.}
  \label{tab:big_step_inference}
\end{table}

\section{Inference Composition}

Consider running a probabilistic program $f$ to draw a sample $x$ from an unnormalized density $\gamma_{f}(x\mid y)$.  We denote
drawing a properly weighted sample $x$ with weight $w$ from $\gamma_{f}(x\mid y)$ via $f$ as
$(x,w) \eval f(y)$.  Since proper weights are ratios of unnormalized densities, any joint distribution formed by a probabilistic program constitutes a proper weight.  We can thus express as a properly weighted sampler any inference technique which only requires producing samples from programs.

Table~\ref{tab:big_step_inference} shows inference rule semantics for several inference combinators in terms of how they take properly weighted samplers as arguments and return properly weighted samplers in turn.  Note that  when $q$ denotes a transition kernel that satisfies detailed balance, as used in Markov chain Monte Carlo methods, the new proper weight $w'=w$.

The proposed combinator framework is a natural fit for modern varational methods for training deep probabilistic models. Because the weight $w$ is an unbiased estimator of the normalizer, we can use its logarithm as an evidence lower bound (ELBO) \cite{kucukelbir2017automatic} or evidence upper bound (EUBO) \cite{bornschein2015reweighted} to perform variational inference by automatic differentiation in PyTorch. For a parameterized density $\gamma_\theta(x \mid y)$, we can approximate the gradient $\nabla_\theta \log Z_\theta(y)$ using the Monte Carlo estimator
\begin{align*}
    % \nabla_\theta \log \hat{Z}_\theta(y)
    % = 
    \sum_{k} \frac{w^{k}}{\sum_l w^{l}}
    \nabla_\theta \log \gamma_{\theta}(x^{k} \mid y).
    %x^{k}, w^{k} \eval \gamma_{\theta}(x \mid y).
\end{align*}
%where $x^{k}, w^{k} \eval \gamma_\theta(x ; y)$ are generated by evaluating the model. 
%By Jensen's inequality $\mathbb{E}[\log \hat{Z}_\theta(y)] \le \log \mathbb{E}[\hat{Z}_\theta(y)]$, so this gradient maximizes a stochastic lower bound \cite{}.
When we sample from an inference model $q_\phi(x \mid y)$ we can perform wake-sleep style inference by minimizing the objective $\text{KL}(\gamma_\theta(x \,;\, y) / Z_\theta(y) \,||\, q_\phi(x \,|\, y))$ using the estimator
\begin{align*}
    -\sum_{k} \frac{w^{k}}{\sum_l w^{l}}
    \nabla_{\phi} \log q_\phi(x^k \mid y).
\end{align*}
Note that the gradient w.r.t.~$\theta$ computes $\nabla_\theta \log w$ whereas the gradient w.r.t.~$\phi$ computes $-\nabla_\phi \log w$. In other words, we can perform variational inference in any properly weighted model by automatic differentiation on the importance weights.

\section{Evaluation}

\begin{figure*}[!t]
\begin{center}
    \includegraphics[width=\textwidth]{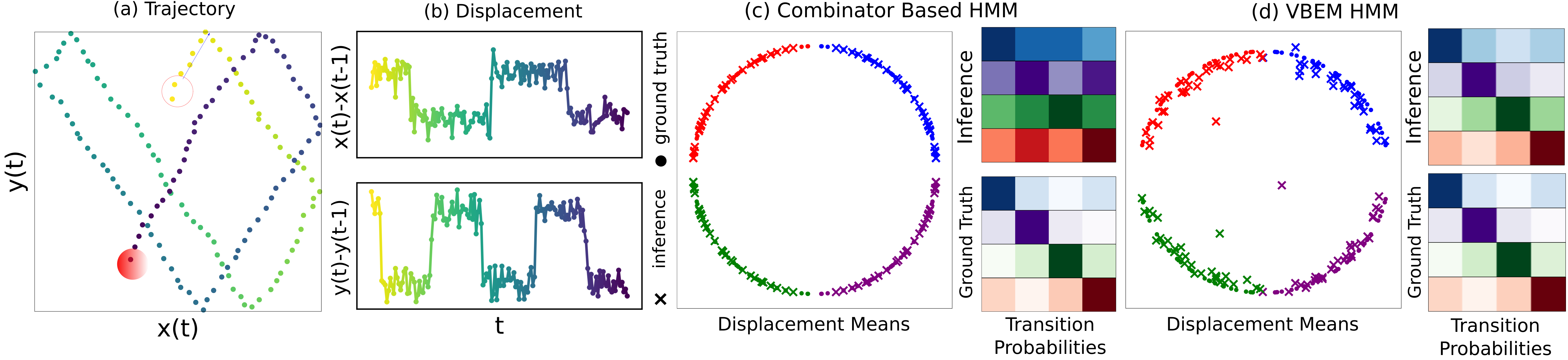}
    % % ~~
    % \includegraphics[height=1.5in]{figures/trajectory_and_velocity/trajectory_1.pdf}
    %  \includegraphics[height=1.5in]{figures/trajectory_and_velocity/velocity_1.pdf}
    % % ~~
    % % \raisebox{12pt}{
    %  \includegraphics[height=1.15in]{figures/baseline/baseline_results_v3.pdf}
    % % ~~
    % % \raisebox{12pt}{
    %  \includegraphics[height=1.15in]{figures/combinators/combinator_results.pdf}
\end{center}
\vspace{-3ex}
\caption{
\label{fig:hmm}
Combinator-based variational inference in hidden Markov models (HMM). a) A bouncing ball trajectory with initial velocity. b) The displacement along x and y axis, respectively. c) Inferred travel directions and transition probabilities from combinator-based wake-sleep Sequential Monte Carlo (SMC). d) Inferred travel directions and transition probabilities from Variational Bayesian Expectation Maximization (VBEM).
}
\end{figure*}

Figure~\ref{fig:hmm} shows inference results on simulated data. The data models a bouncing particle trajectory in a closed box (Fig.~\ref{fig:hmm}a). This trajectory has a piece-wise constant noisy velocity, which means that the displacements at each time step (Fig.~\ref{fig:hmm}b) can be described by an HMM with Gaussian observations, where each state's observation mean corresponds to the average velocity along one of four possible directions of motion. We compare wake-sleep SMC inference results for a combinator-based implementation (Fig.~\ref{fig:hmm}c) to those obtained using variational Bayesian expectation maximization (VBEM) (Fig.~\ref{fig:hmm}d), for a set of 30 time series that each contain 200 time points.  VBEM optimizes the exclusive Kullback-Leibler (KL) divergence, $\mathcal{D}_{KL}(q\mid\mid p)$, while in our combinator-based inference we optimized $\mathcal{D}_{KL}(p\mid\mid q)$, approximating the posterior with greater variance.  The combinator-based HMM implementation required 64 lines of model specific code.

\section{Extensions and Future Work}

In this work, we have considered the fewest assumptions possible about the structural form of the generative model, and so we believe these inference techniques to be the most suited to dynamic probabilistic programs with unbounded dimensionality. Our next step will be to apply combinators to modeling intuitive physics in perception.  A variety of extensions are possible that make use of some information about the graphical structure of a model. One opportunity is to apply enumeration or belief propagation to components of the model that are amenable to such operations. Recent work by the Pyro team on sum-product implementations for deep probabilistic programs is relevant in this context \cite{uber2018ubersum}. Recent just-in-time compilation strategies \cite{pytorch2018script} could be adapted to construct static graphs for such models, which an optional API could expose to implement inference optimizations.

\subsubsection*{Acknowledgments}

The authors would like to thank David Tolpin for his early interest in this line of work, and the two anonymous
reviewers at the Bayesian Nonparametrics workshop for their detailed feedback.

\medskip
\small

\bibliographystyle{plain}
\bibliography{nips_2018}

\end{document}